\documentclass[letterpaper, 10 pt, conference]{ieeeconf} 

\IEEEoverridecommandlockouts                              

\usepackage{amsmath,amsfonts}
\usepackage{algpseudocode,algorithm,algorithmicx}
\usepackage{array}
\usepackage[caption=false,font=normalsize,labelfont=sf,textfont=sf]{subfig}
\usepackage{textcomp}
\usepackage{stfloats}
\usepackage{url}
\usepackage{verbatim}
\usepackage{graphicx}
\usepackage{cite}
\usepackage{hyperref}
\hypersetup{
    colorlinks=true,
    linkcolor=black,
    filecolor=black,      
    urlcolor=blue,
    pdftitle={Overleaf Example},
    pdfpagemode=FullScreen,
    }

\begin{document}

\title{\LARGE \bf Adaptive Goal Management System of Robots}

\author{Muhammad Kazim$^{1}$, Michael Muldoon$^{2}$ and Kwang-Ki K. Kim$^{1,*}$
\thanks{$^{1}$M. Kazim and K.-K. K. Kim are with the Department of Electrical and Computer Engineering at Inha University, Incheon, Republic of Korea. (Email: {\tt kazim@inha.ac.kr}; {\tt kwangki.kim@inha.ac.kr})}
\thanks{$^{2}$M. Muldoon is with Command Robotics, 10230 Rue Panis-Charles, Montreal, Quebec, Canada. (Email: {\tt michael.muldoon.home@gmail.com})}
\thanks{$^*$Corresponding author: K.-K. K. Kim}
}

\maketitle

\begin{abstract}
This paper considers the problem of managing single or multiple robots and proposes a cloud-based robot fleet manager, \emph{Adaptive Goal Management (AGM)} System, for teams of unmanned mobile robots. The AGM system uses an adaptive goal execution approach and provides a restful API for communication between single or multiple robots, enabling real-time monitoring and control. The overarching goal of AGM is to coordinate single or multiple robots to productively complete tasks in an environment. There are some existing works that provide various solutions for managing single or multiple robots, but the proposed AGM system is designed to be adaptable and scalable, making it suitable for managing multiple heterogeneous robots in diverse environments with dynamic changes. The proposed AGM system presents a versatile and efficient solution for managing single or multiple robots across multiple industries, such as healthcare, agriculture, airports, manufacturing, and logistics. By enhancing the capabilities of these robots and enabling seamless task execution, the AGM system offers a powerful tool for facilitating complex operations. The effectiveness of the proposed AGM system is demonstrated through simulation experiments in diverse environments using {\tt ROS1} with {\tt Gazebo}. The results show that the AGM system efficiently manages the allocated tasks and missions. Tests conducted in the manufacturing industry have shown promising results in task and mission management for both a single Mobile Industrial Robot and multiple Turtlebot3 robots. To provide further insights, a supplementary video showcasing the experiments can be found at~\url{https://github.com/mukmalone/AdaptiveGoalManagement}. 
\end{abstract}



\section{Introduction}\label{sec:intro}
Robotic swarms, comprising multiple robots, are increasingly being deployed in various industries to perform complex tasks efficiently \cite{williams2020utilizing,issa2019survey}. However, managing a swarm of robots poses significant challenges, including task allocation, coordination, and communication. To address these challenges, an adaptive goal management (AGM) system is proposed, which acts as a cloud-based robot fleet manager for unmanned ground vehicles (UGVs), unmanned aerial vehicles (UAVs), and unmanned surface vehicles (USVs). The AGM system uses an adaptive goal execution approach, enabling efficient and effective task allocation and execution.

The proposed AGM system is designed to provide a versatile solution for managing single or swarms of robots across various industries, including healthcare, agriculture, airports, manufacturing, and logistics. The system's restful API allows for real-time communication between a single or swarm of robots, enabling seamless monitoring and control. The AGM system's adaptability and scalability make it an ideal solution for managing single or swarms of robots in diverse environments.

Previous works have proposed various solutions for managing a single or swarm of robots, including behavior-based approaches \cite{xu2014behavior}, hierarchical control structures \cite{jiamin2023hierarchical}, and decentralized control architectures \cite{testa2021choirbot,metoui2020path}. However, these methods often lack adaptability and scalability, making them unsuitable for large-scale robot fleets. In contrast, the proposed AGM system is designed to be adaptable and scalable, making it suitable for managing large swarms of robots in diverse environments.

To validate the proposed AGM system's effectiveness, simulation experiments were conducted in different environment scenarios, using {\tt ROS1/GAZEBO} environment. The results show that the AGM system efficiently manages the allocated tasks and missions. Furthermore, tests conducted on single MIR \cite{MIR} and multiple Turtlebot3 \cite{Turtlebot3} also indicate promising results in task and mission management. The supplementary materials are available at \url{https://www.command-central.com/} and \url{http://www.commandrobotics.ai/}. In this paper, AGM and Command-Central are used interchangeably in some of the diagrams.

In conclusion, the proposed AGM system offers an adaptable and efficient solution for managing single or swarms of robots across various industries, enhancing their capabilities and facilitating complex tasks. The system's restful API enables real-time communication between the robots, providing seamless monitoring and control. The results of the simulation experiments on MIR and Turtlebot3 tests demonstrate the AGM system's effectiveness in managing tasks and missions.

\subsection*{Related Work}
Managing a single or swarm of robots in different environments is a challenging task that has gained significant attention from researchers in recent years. Recently several approaches have been proposed to address the challenges of managing single or robotic swarms. Recently, Testa A. et al.~\cite{testa2021choirbot} proposed ChoiRbot that is a ROS2-based toolbox for distributed cooperative robotics. It provides a fully-functional toolset to execute complete distributed multi-robot tasks, with a particular focus on networks of heterogeneous robots without a central coordinator.  

In \cite{2010Centralized,Centralized2018,centralized2021overview}, used central control systems, where a single controller manages the swarm of robots. However, this approach is not scalable, and a single point of failure can affect the entire swarm. To address this limitation, distributed control systems have been proposed in \cite{distributed2020,distributed2022b,distributed2021}, where the swarm of robots communicates with each other and makes decisions based on local information. Another approach is the use of decentralized control systems \cite{decentralized2021,testa2021choirbot,decentralized2022}, where each robot in the swarm acts autonomously, making decisions based on local information. This approach provides scalability, fault tolerance, and adaptability, but it can lead to coordination problems and reduced efficiency.

To address the limitations of the above approaches, several researchers in \cite{adaptive2021,adaptive2022} have proposed using decentralized-based adaptive control systems that can adapt to changing environments and tasks. These systems use feedback from the swarm of robots and the environment to adjust their behavior and decision-making processes. In recent years, in \cite{DRL2020,DRL2020b}, there has been a significant focus on using machine learning and artificial intelligence techniques to manage robotic swarms. These techniques enable the swarm of robots to learn from experience and improve their decision-making processes over time.

In conclusion, several approaches have been proposed to manage robotic swarms, including central control systems, distributed control systems, decentralized control systems, and adaptive control systems. The use of machine learning and artificial intelligence techniques has also gained significant attention from researchers. Each approach has its strengths and limitations, and the choice of approach depends on the specific application and task at hand. The proposed AGM system uses an adaptive goal execution approach and restful API communication to address the challenges of managing single and large-scale robotic swarms in various industries.

\subsection*{Problem Statement}
The deployment of robotic swarms in various industries, including healthcare, agriculture, airports, manufacturing, and logistics, has become increasingly popular. These swarms consist of multiple robots that work together to perform complex tasks efficiently towards potentially a larger common goal. However, managing these swarms poses significant challenges, including task allocation, coordination, and communication.

One of the primary challenges in managing robotic swarms is the efficient allocation of tasks. The allocation of tasks to the individual robots must be done in such a way that the overall objective of the swarm is achieved optimally. Traditional task allocation algorithms do not account for the dynamic changes in the swarm's environment and the robots' capabilities, leading to sub-optimal performance, bottlenecks, and deadlocks.

Coordination is another challenge in managing robotic swarms. Coordination involves ensuring that the individual robots in the swarm work together to achieve the overall objective efficiently. Traditional coordination approaches, such as centralized control, are not scalable and cannot account for the dynamic changes in the swarm's environment.

Finally, communication is a critical challenge in managing robotic swarms. The individual robots in the swarm must communicate with each other to ensure efficient task allocation and coordination. Traditional communication approaches are often limited by the number of robots in the swarm, making them unsuitable for large-scale robotic swarms.


\subsection*{Contribution}
Traditional approaches to managing single or robotic swarms have limitations, making them unsuitable for large-scale robotic swarms. A new approach was required to overcome the research gap presented in the related works and address the challenges presented in the problem statement. We proposed an adaptive goal management (AGM) system as a cloud-based robot fleet manager for UGVs, UAVs, and USVs. The AGM system uses an adaptive goal execution approach, enabling efficient and effective task allocation and execution. The system's restful API allows for real-time communication between a single or swarm of robots, enabling seamless monitoring and control. The AGM system's adaptability and scalability make it an ideal solution for managing single or large swarms of robots in diverse environments.

The proposed AGM system is designed to be adaptable and scalable, making it suitable for managing single or multiple robots in diverse environments. It presents a versatile and efficient solution for managing single or multiple robots across multiple industries, such as healthcare, agriculture, airports, manufacturing, and logistics. For example, in healthcare, robotic swarms can be deployed to transport medicine and equipment, reducing the workload on healthcare staff. In agriculture, robotic swarms can be used to perform tasks such as planting and harvesting crops, increasing productivity and efficiency. In manufacturing and logistics, robotic swarms can be deployed to perform tasks such as inventory management and order fulfillment, reducing costs and improving efficiency.

In conclusion, the proposed AGM system's objective is to provide an efficient and adaptable solution for managing robotic swarms in various industries. The motivation behind the proposed system is to enhance the capabilities of robotic swarms and facilitate complex tasks efficiently, addressing the challenges of managing large-scale robotic swarms. The system's restful API enables real-time communication between the swarm of robots, providing seamless monitoring and control, and has significant implications for various industries.

\section{Architecture Description}\label{sec:method}
This section provides an overview of the high-level architecture of the AGM system. Firstly, we present a broad description of the software architecture, followed by a detailed breakdown of each component.

\subsection{Overview of the Software}\label{sec:method:1}
The Adaptive Goal Management (AGM) system is a cloud-based system that utilizes secure socket messaging and the {\tt https} protocol.  This architecture is designed to take advantage of the tremendous power, speed, and cost-effectiveness of modern cloud systems, which have become increasingly robust and secure. Any robot with an internet connection can become part of the fleet of robots being managed by the AGM system. By adhering to the {\tt https API} protocol, robots can quickly and efficiently request and receive their tasks, making the system incredibly versatile and easy to use. Additionally, the secure socket messaging employed by the AGM system ensures that all communication is fully protected, guaranteeing the safety and integrity of all data exchanged between robots and the cloud-based system.

The AGM system software itself has three layers to the application. The webserver is written in JavaScript using the {\tt Node.js} framework.  The User Interface {\tt (UI)} also is written in JavaScript using the React framework.  The database uses the {\tt MongoDB} database, which is a cross-platform document-oriented database program classified as a {\tt NoSQL} database program. This multi-layered software architecture offers a robust and scalable solution that is well-suited for managing robots efficiently and securely.

The AGM system's browser-based {\tt UI} allows for easy human interaction with the robot fleet from any device, facilitating task assignment and collaboration between humans and robots in real-time. Robots and automated workstations communicate with the AGM system using a {\tt RESTful API}, enabling seamless and efficient data exchange between the systems. The communication node between the AGM system and robots/workstations, called {\tt agm\_comm}, is written in C++ and is available as open-source software.  

\subsection{Network Architecture}\label{sec:method:2}
\begin{figure}[t]
	\centering
	\includegraphics[width=1.0\linewidth]{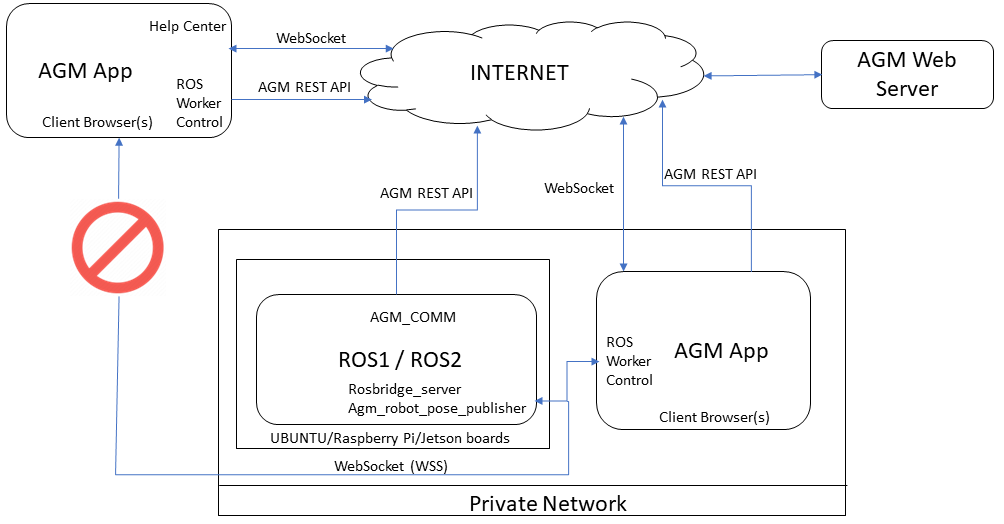}\vspace{-2mm}
	\caption{Network Architecture}
	\label{Architecture}
\end{figure}

The proposed network architecture of the AGM system is illustrated in Fig.~\ref{Architecture}. Typically, robots are situated within a private network to minimize external interference. To communicate with the AGM system, robots initiate communication using the {\tt https} protocol. This communication originates from a trusted source within the private network and travels to the cloud, where security measures are implemented to prevent unauthorized access by external actors seeking to breach the private network's security.

The private network provides a secure environment for the robots to operate without the risk of external interference. Through the use of sockets and the {\tt ROS} architecture, robots can be controlled from a web browser by running the {\tt rosbridge} server and robot pose publisher nodes. This allows for seamless communication between the robots and the AGM system, enabling efficient task management and collaboration. However, to ensure safety and maintain network security, control from outside the private network is not permitted. By implementing these measures, the AGM system ensures the safety and security of the robots while providing a reliable and efficient means of communication. The AGM system provides a browser-based {\tt UI} that can be accessed from any device with internet access, allowing users to monitor and trigger tasks and set up new functionality. The user-friendly interface enables dynamic and agile management of the robots and provides real-time feedback on task status to quickly address any issues.

\subsection{Structure of the DATA}\label{sec:method:3}
\begin{figure}[t]
	\centering
	\includegraphics[width=1.0\linewidth, height= 5.0cm]{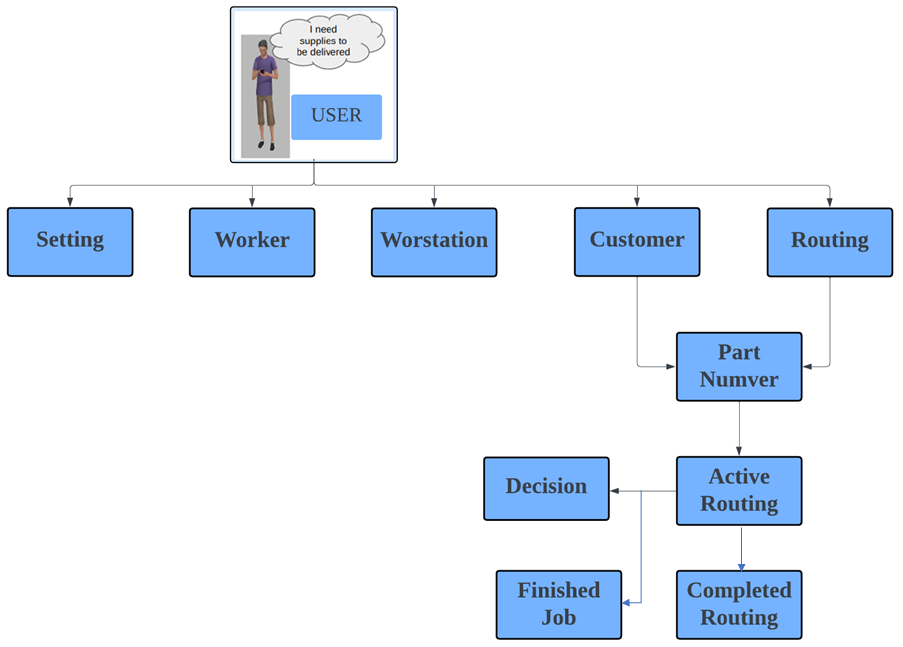}\vspace{-2mm}
	\caption{Structure of the Data}
	\label{Structure of Data}
\end{figure}

The document structure of the {\tt MongoDB} database is depicted in Fig.~\ref{Structure of Data}. The hierarchy starts with the user, who can create an unlimited number of robots, tasks, and workstations. The AGM system supports authentication, which enables it to run multiple organizations with any number of robots. Routing is a set of instructions that defines a sequence of tasks required to achieve a larger goal, such as the production of a particular product. When a user activates a routing, the associated set of instructions becomes part of the available work that can be performed and assigned to a robot. Workers, which can be robots, humans, or other equipment capable of performing work, execute the assigned tasks. A workstation is where an operation is performed by something other than the worker. The AGM system keeps track of each decision made, such as task assignment, worker activity, and completed routings.

\subsection{API Payloads}\label{sec:method:4}
\begin{figure}[t]
	\centering
	\includegraphics[width=1.0\linewidth]{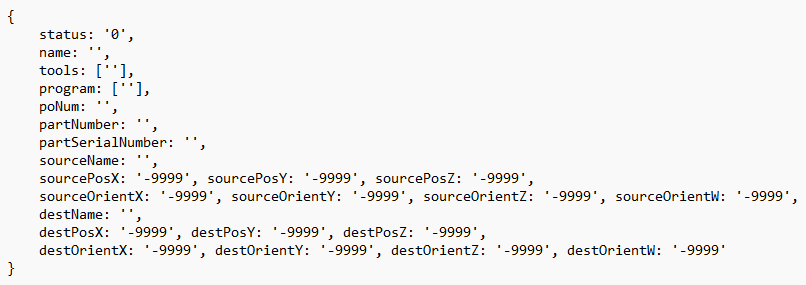}\vspace{-2mm}
	\caption{API Payloads}
	\label{API payloads}
\end{figure}

Robots and Workstations interact with the AGM system using the Restful API as shown in Fig. \ref{API payloads} via the {\tt HTTPS} protocol. To query the AGM system for the next task, a {\tt curl} request can be used, specifying the {\tt URL} with the appropriate {\tt robot\_ID} in the key parameter. For example, the {\tt curl} request to retrieve the next task for a robot with the ID "{\tt robot\_ID}" would be: "curl: \url{https://www.command-central.com/workerGetNextJob?key=your-robot-id-goes-here"}. This API allows for seamless communication between the robots and the AGM system, enabling efficient task management and collaboration while ensuring the security and privacy of the communication.

After a job is assigned to a robot, AGM sends a payload in response. The payload may contain various pieces of information depending on the robot's implementation. At a minimum, the payload includes a "status" field, which indicates whether the response is valid (status=1) or not (status=0). Additionally, the payload contains the source and destination coordinates, which are formatted in 3D space to allow drones and 3D robots to utilize the task system. These coordinates enable the robot to perform the required task accurately and efficiently.

\subsection{Task prioritization Algorithms}\label{sec:method:5}
\begin{figure}[t]
	\centering
	\includegraphics[width=.95\linewidth, height= 7.0cm]{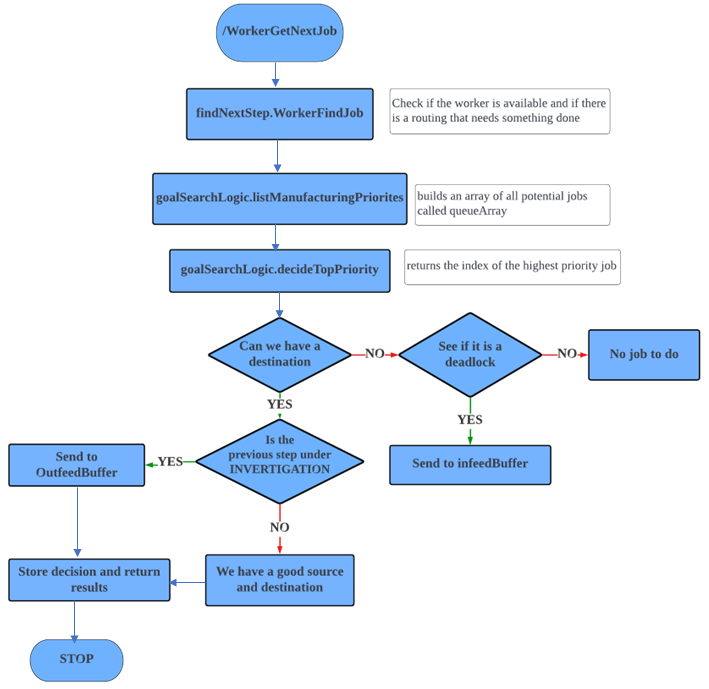}\vspace{-2mm}
	\caption{Task prioritization algorithm}
	\label{task algorithm}
\end{figure}

Task prioritization algorithms are essential for the efficient and effective management of single or multiple robots. With many active routings in the system and dynamic states of workstations, it is crucial to decide which task should be performed next. The AGM system uses a {\tt Restful API} to receive requests from robots regarding their next task. Once a robot becomes available for the next task, AGM works through a decision tree shown in Fig. \ref{task algorithm}, which outlines the task prioritization algorithms, to select the most appropriate task for that robot to perform. The decision tree takes into consideration various factors, such as the priority of the task, the type of robot, the distance to the task location, and the availability of resources. Once a task is selected, it is removed from the list of available tasks, ensuring that another robot does not request the same task. The task prioritization algorithms enable AGM to optimize the allocation of tasks and ensure that robots are utilized efficiently, leading to better productivity and reduced operational costs.

\subsection{Notifications and Interventions}\label{sec:method:6}
\begin{figure}[t]
	\centering
	\includegraphics[width=.95\linewidth, height= 9.0cm]{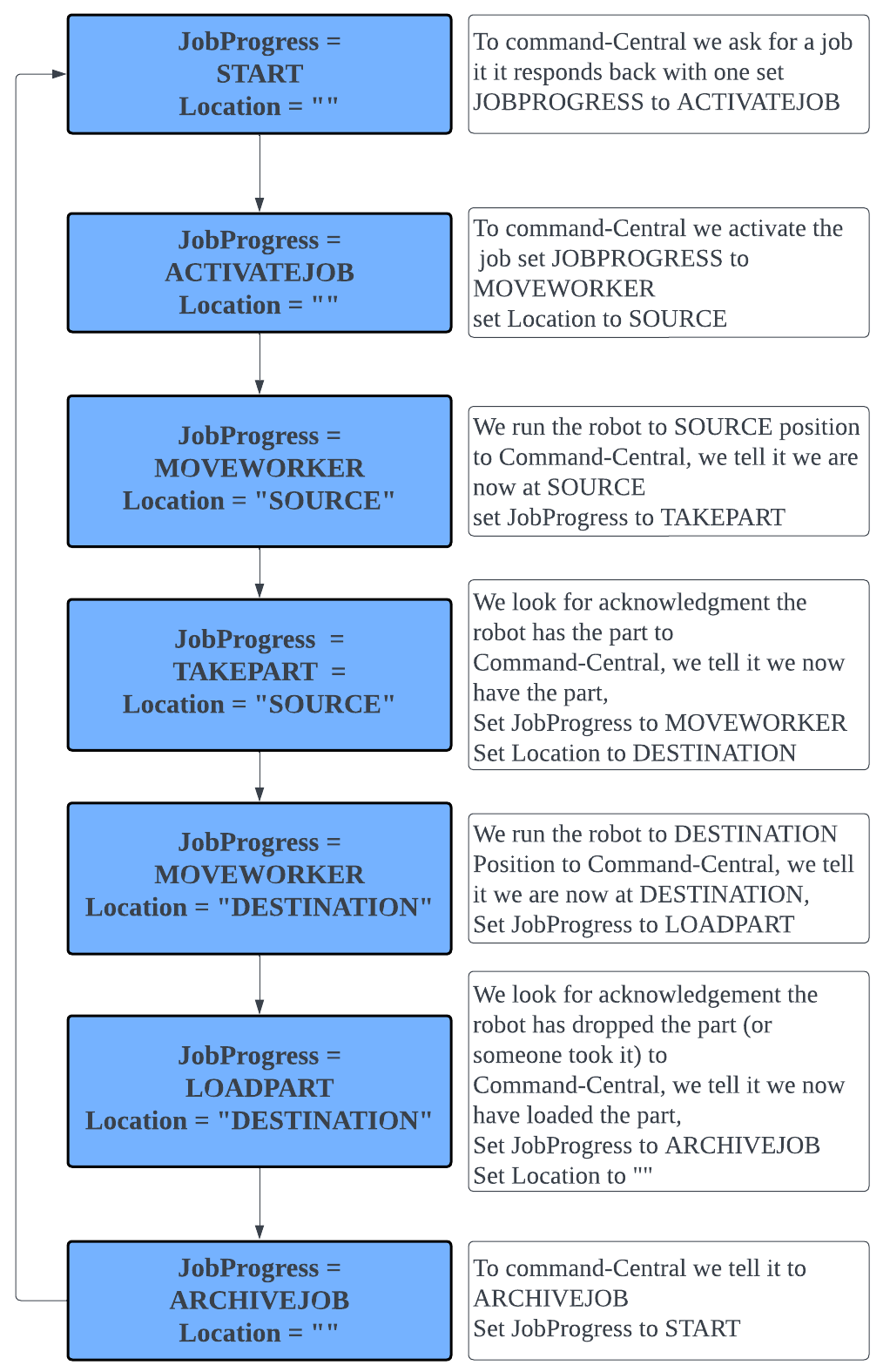}\vspace{-2mm}
	\caption{Robots working sequence}
	\label{Robots sequence}
\end{figure}

When a robot starts working on a task assigned by AGM, it periodically informs the AGM system about its progress through the {\tt Restful API}. The workflow diagram (referring to Fig. \ref{Robots sequence}) illustrates the sequence of messages exchanged between the robot and AGM. These messages update the user interface (UI) to show the current progress of the task. Additionally, the robot also informs AGM about its current position and battery level, which can be utilized during the decision-making process for task prioritization. This information allows AGM to make informed decisions about which tasks to assign to the robots and when ensuring efficient and effective task execution.

\subsection{Customizable and Extensible Web Interface}\label{sec:method:7}
The AGM system's web interface is highly customizable and extensible, providing users with the flexibility to adapt the system to their specific needs. The interface allows users to create and manage their own custom robots, tasks, and environments. Users can add, modify or delete existing robots, tasks, and environments to suit their requirements. Additionally, the web interface provides real-time updates on the status of each robot, the current task it is performing, and the progress of that task. This information is displayed through an interactive dashboard that is easy to understand and navigate. The customizable and extensible web interface of the AGM system allows users to configure the system to suit their specific needs, making it a highly versatile and efficient solution for managing single or multiple robots across multiple industries.

\subsection{Command and Control in real-time}\label{sec:method:8}
The AGM system provides real-time command and control capabilities for managing single or multiple robots. The system uses a {\tt RESTful API} to enable communication between the robots and the AGM server, allowing for seamless monitoring and control. The real-time command and control feature enables the AGM system to adapt dynamically to changes in the environment and adjust the robot's actions accordingly. This feature allows for agile task execution, ensuring that the robots are always performing the most important tasks in the most efficient manner. With real-time command and control, the AGM system offers a powerful tool for managing complex operations in various industries, including healthcare, agriculture, manufacturing, and logistics.





\section{Simulations and Discussions}\label{sec:sim}
In the simulation, two case studies are presented; the first case study involves a single Mobile Industrial Robot (MIR) , while the second case study uses multiple Turtlebot3 in the manufacturing industry. These case studies are aimed at demonstrating the effectiveness of the AGM system in managing tasks for different fleets in different environments. The experiments are conducted using {\tt ROS1/GAZEBO}, and the results show that the AGM system efficiently manages the allocated tasks and missions. By showcasing the experiments, the proposed AGM system offers further insights into the adaptability and scalability of the system and its ability to handle multiple robots in different scenarios. 

\begin{figure}[t]
	\centering
	\includegraphics[width=.95\linewidth]{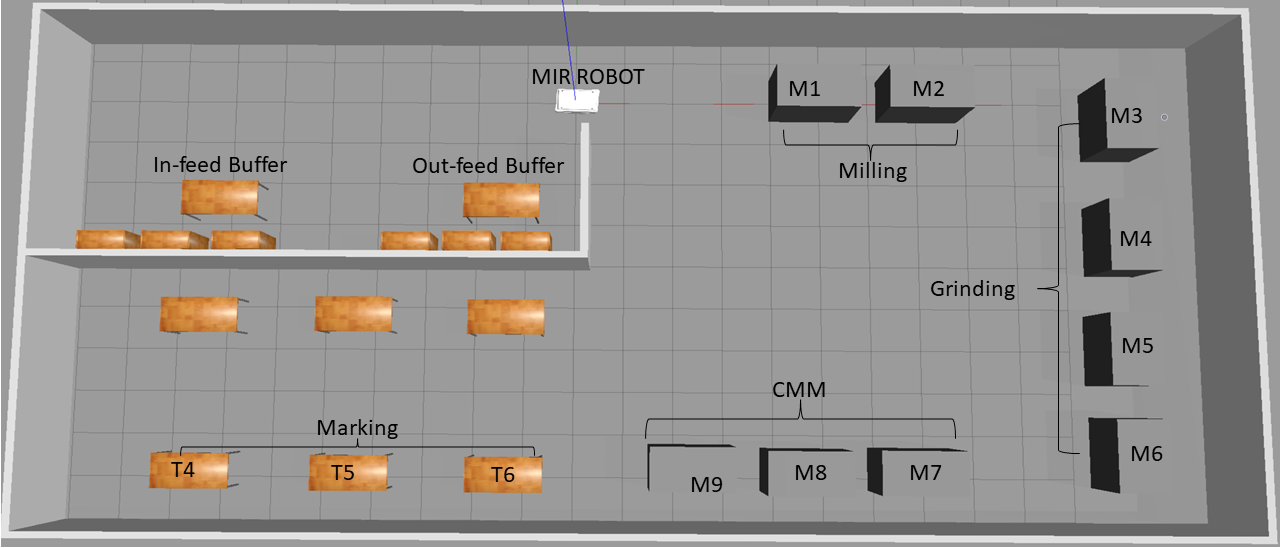}\vspace{-2mm}
	\caption{Single MIR Robot in the Manufacturing Industry}
	\label{fig:single_MIR}
\vspace{2mm}
	\centering
	\includegraphics[width=.95\linewidth]{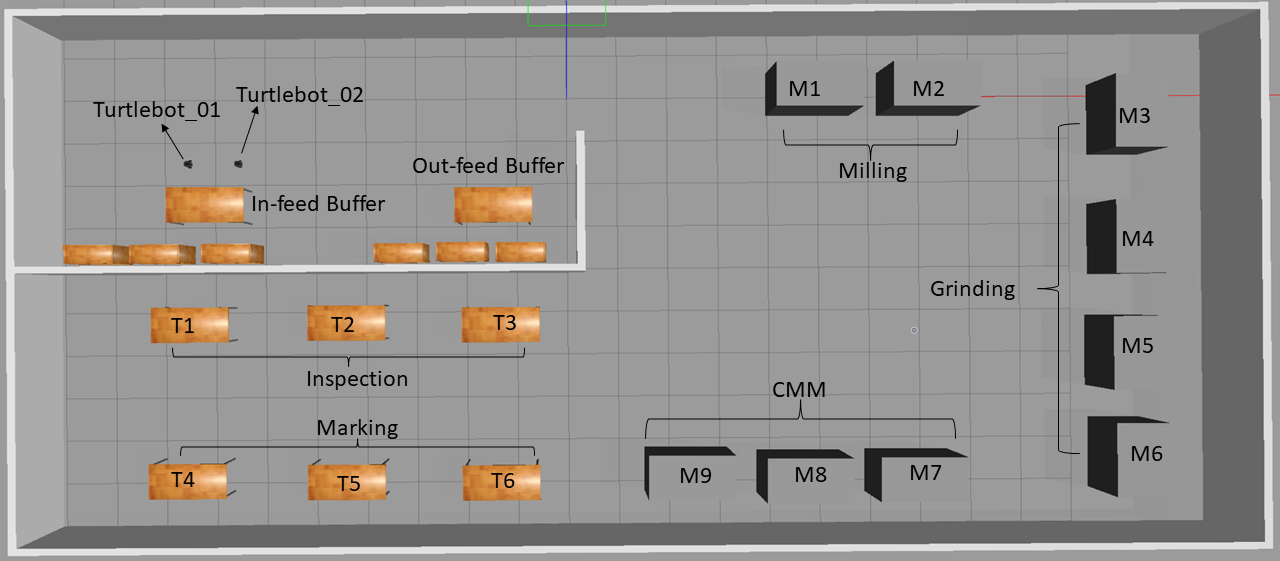}\vspace{-2mm}
	\caption{Multiple Turtlebot3 Robot in the Manufacturing Industry}
	\label{fig:Multiple_Turtlebot3}
\end{figure}

First, we define the setup of workstations in a manufacturing environment for optimizing production efficiency. The configuration includes a single in-feed Buffer, two milling stations, four grinding stations, three CMM stations, and three marking stations for the single MIR robot shown in Fig. \ref{fig:single_MIR} and for multiple Turtlebot3 are shown in Fig. \ref{fig:Multiple_Turtlebot3}. {\tt RVIZ} is used to accurately get the position and orientation of the stations, and routing configurations are designed based on product requirements. Customer profiles and unique part numbers are also established to enhance efficiency and customer satisfaction. This framework provides an effective approach to optimizing manufacturing processes.   

\begin{figure}[t]
	\centering
	\includegraphics[width=.95\linewidth]{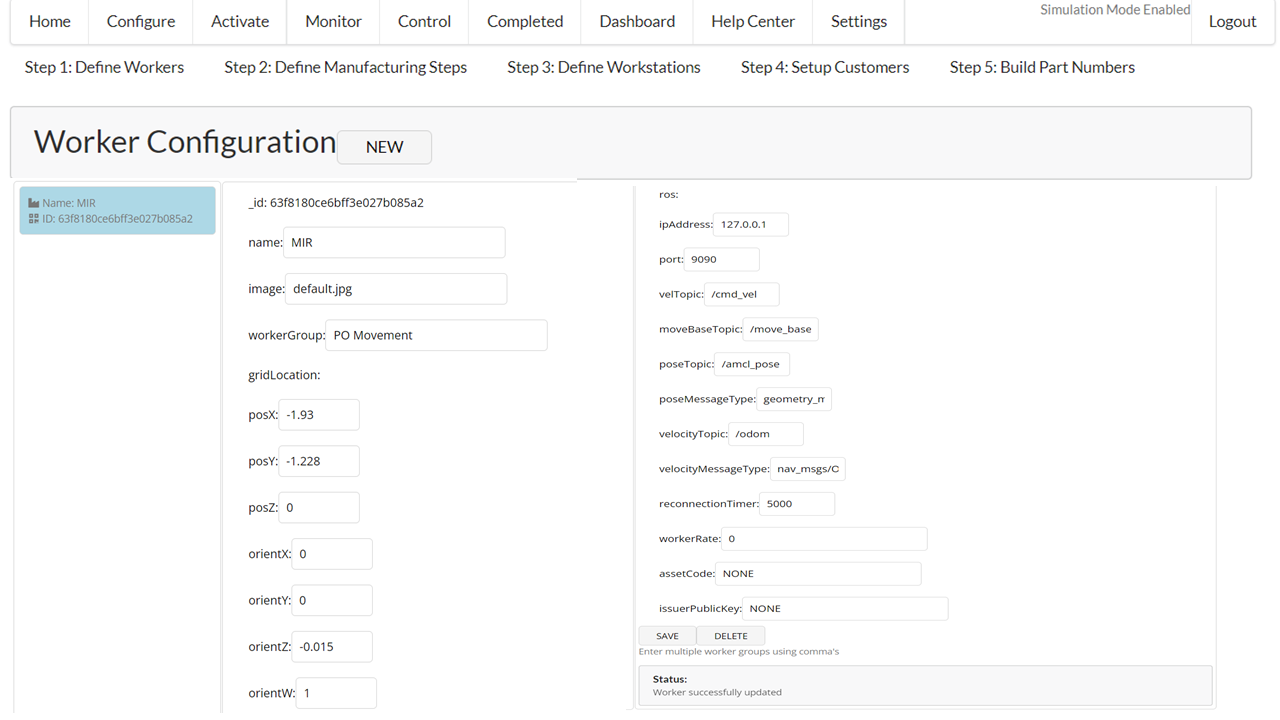}
	\caption{Single MIR Robot Configuration}
	\label{fig:single_MIR_config}
\end{figure}

\begin{figure}[t]
	\centering
	\includegraphics[width=.95\linewidth]{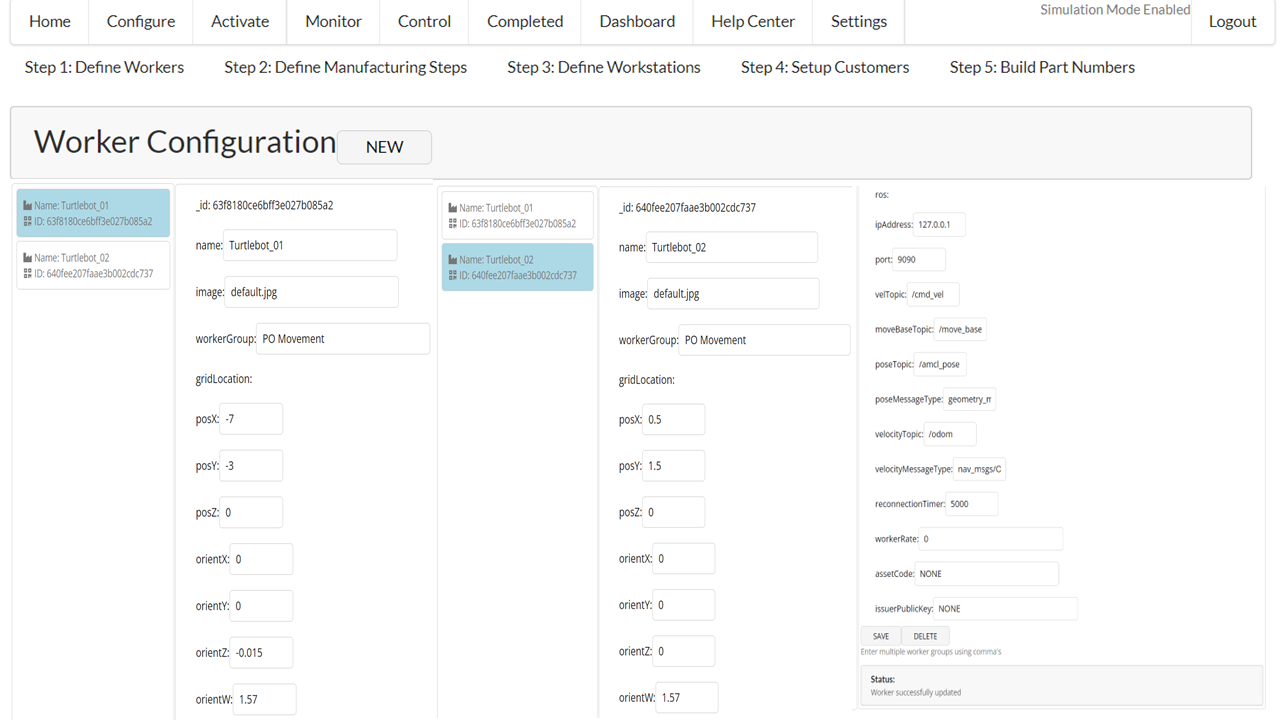}
	\caption{Multiple Turtlebot3 Robot Configuration}
	\label{fig:Multiple_Turtlebot3_config}
\end{figure}

To configure workers in the AGM app, the first step is to define their name, such as "MIR\_robot" for a single robot or "turtlebot\_01" and "turtlebot\_02" for multiple robots, and set the WorkerGroup as PO Movement. The IP address and port numbers should be verified, but other parameters do not require modification. The worker configuration is shown in Fig. \ref{fig:single_MIR_config} for MIR\_robot and Fig. \ref{fig:Multiple_Turtlebot3_config} for multiple robots. After saving the worker configuration, a unique ID is generated and added to the agm\_comm launch file to establish communication between the robots and the AGM app. This communication is critical for the successful coordination and movement of the robots between stations defined in the workstation configuration and for accessing product and customer information.

\begin{figure}[t]
	\centering
	\includegraphics[width=.95\linewidth]{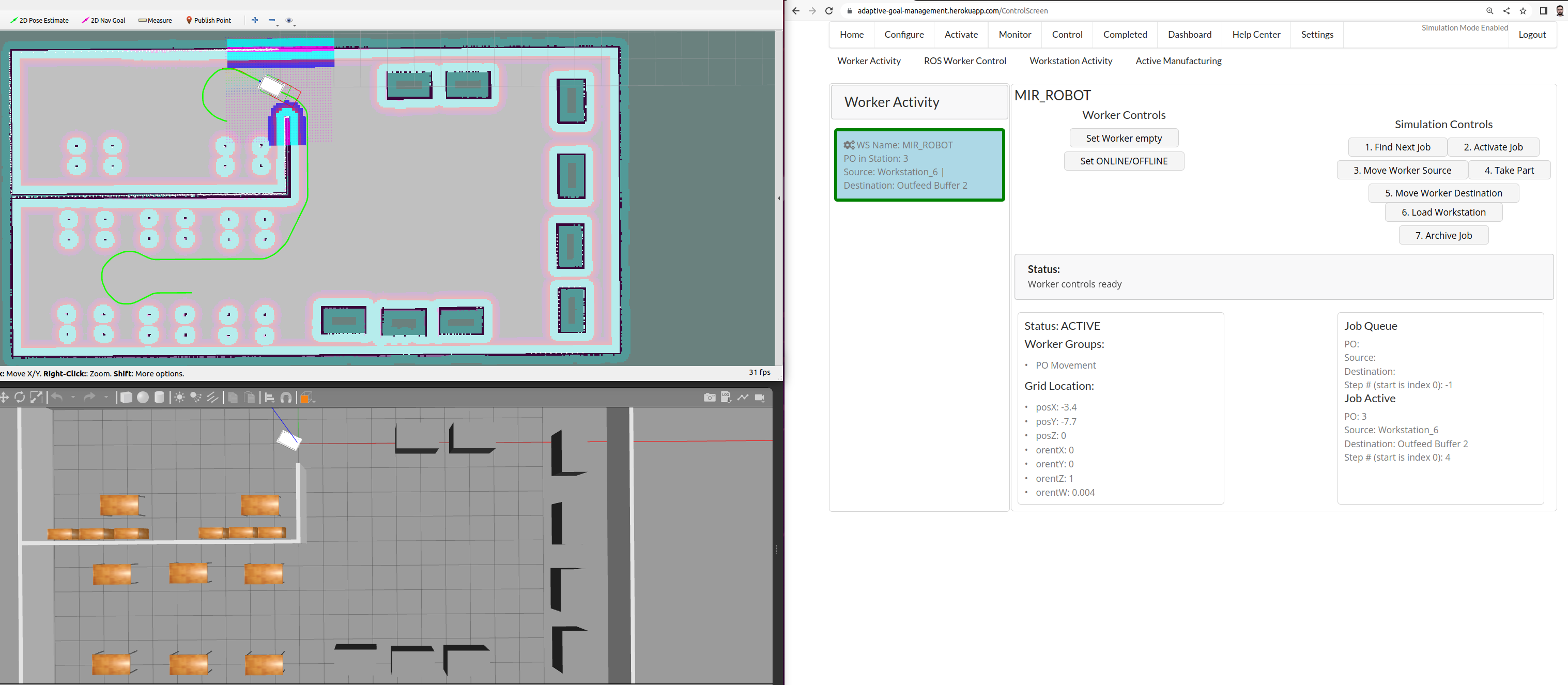}\vspace{-2mm}
	\caption{Single MIR Robot simulation environment, top left is the RVIZ, bottom left is GAZEBO environment, and the right one is showing performance in AGM app}
	\label{fig:MIR_simulation}
\end{figure}

\begin{figure}[t]
	\centering
	\includegraphics[width=.95\linewidth]{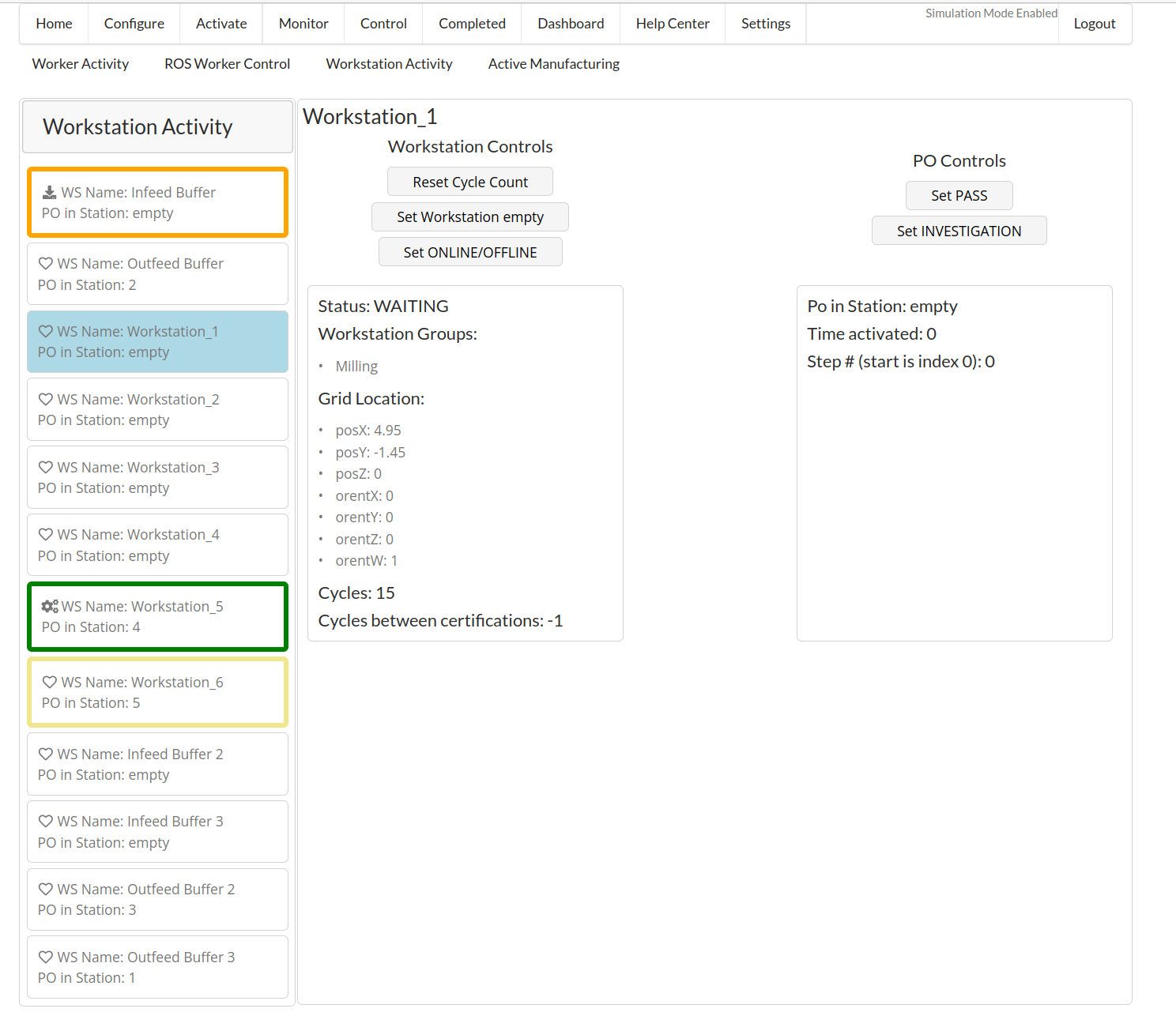}\vspace{-2mm}
	\caption{Single MIR Robot workstation activity}
	\label{fig:MIR_activity}
\end{figure}

\begin{figure}[t]
	\centering
	\includegraphics[width=.95\linewidth]{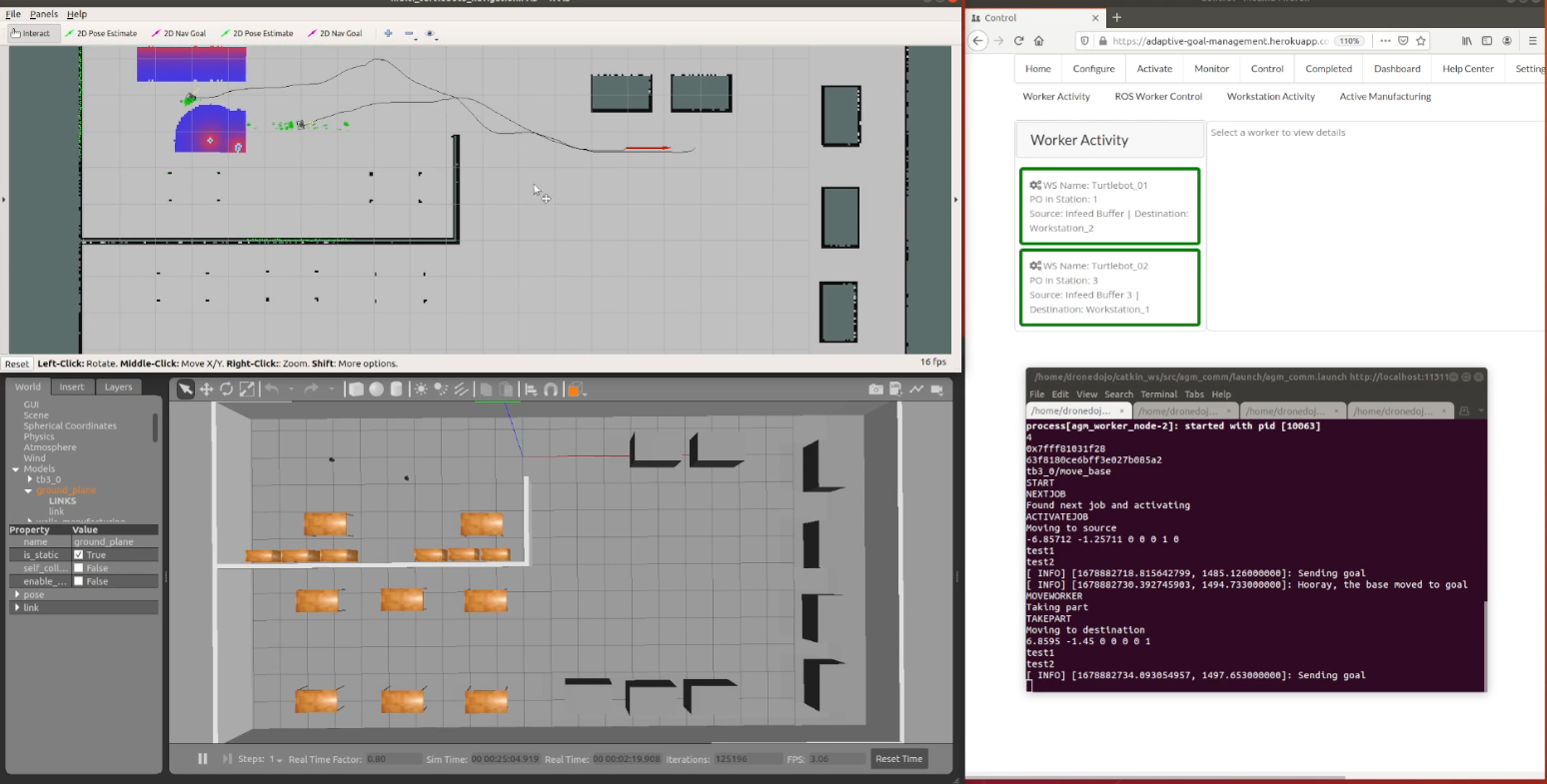}\vspace{-2mm}
	\caption{Multiple Turtlebot3 Robot simulation environment, top left is the RVIZ, bottom left is GAZEBO environment and the right one is showing performance in AGM app}
	\label{fig:Multiple_Turtlebot3_simulation}
\end{figure}

\begin{figure}[t]
	\centering
	\includegraphics[width=.95\linewidth]{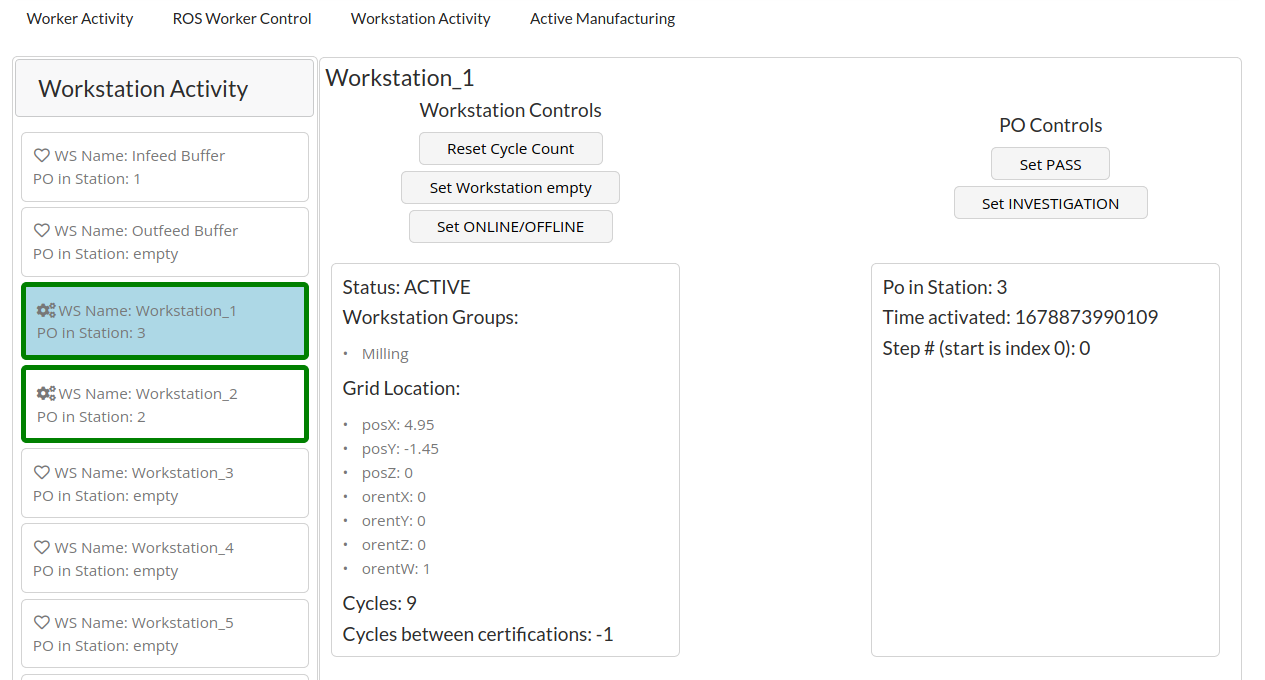}\vspace{-2mm}
	\caption{Multiple Turtlebot3 Robot workstation activity}
	\label{fig:Multiple_Turtlebot3_activity}
\end{figure}

The simulation environment for a single robot, "MIR\_robot", can be seen in Fig. \ref{fig:MIR_simulation}, while the simulation environment for multiple robots, "turtlebot\_01" and "turtlebot\_02", is displayed in Fig. \ref{fig:Multiple_Turtlebot3_simulation}. In the worker activity section of the AGM app, the green sign indicates that communication has been established between the AGM and "MIR\_robot" and a successful task assignment, and the two green signs indicate that the two turtlebot3 robots have successfully established communication with the AGM and recevied their tasks. Workstation activity can be monitored for a single robot in Fig. \ref{fig:MIR_activity}, and for multiple robots in Fig. \ref{fig:Multiple_Turtlebot3_activity}. This information is essential for tracking the progress of robots and ensuring the successful completion of manufacturing tasks.

\begin{figure}[t]
	\centering
	\includegraphics[width=.95\linewidth]{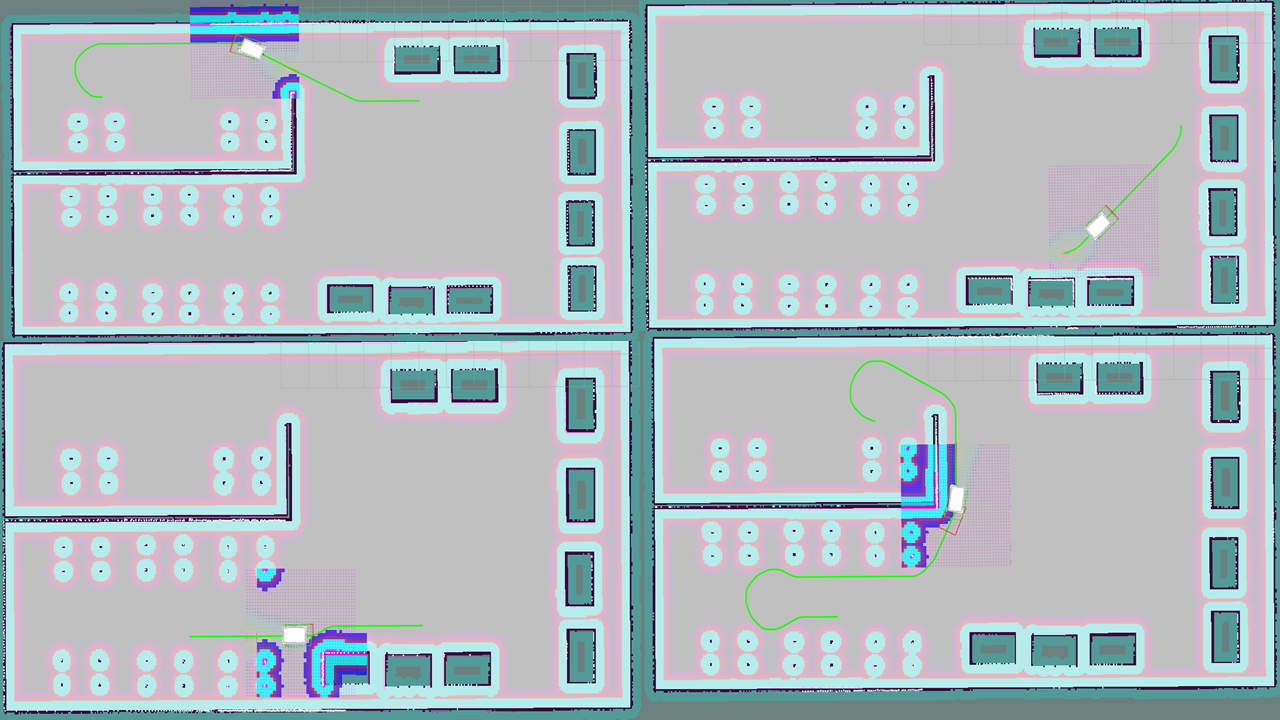}\vspace{-2mm}
	\caption{Single MIR Robot performing different tasks}
	\label{fig:MIR_tasks}
\end{figure}

\paragraph*{Simulation with single MIR robot}
Fig. \ref{fig:MIR_tasks} illustrates the tasks performed by the MIR\_robot. In the top left, the robot moves from the infeed buffer to the Milling workstation, and in the top right, it moves from the Grinding station to the CMM station. The bottom left depicts the robot's movement from the CMM station to the Marking Station, and the bottom right shows the robot delivering the finished product to the outfeed buffer after completing the task at the Marking station. The simulation results with a single MIR robot using the adaptive goal management system are shown in the video\footnote{\url{https://youtu.be/Mg7pJbGGvTo}}. 

\begin{figure}[t]
	\centering
	\includegraphics[width=.95\linewidth]{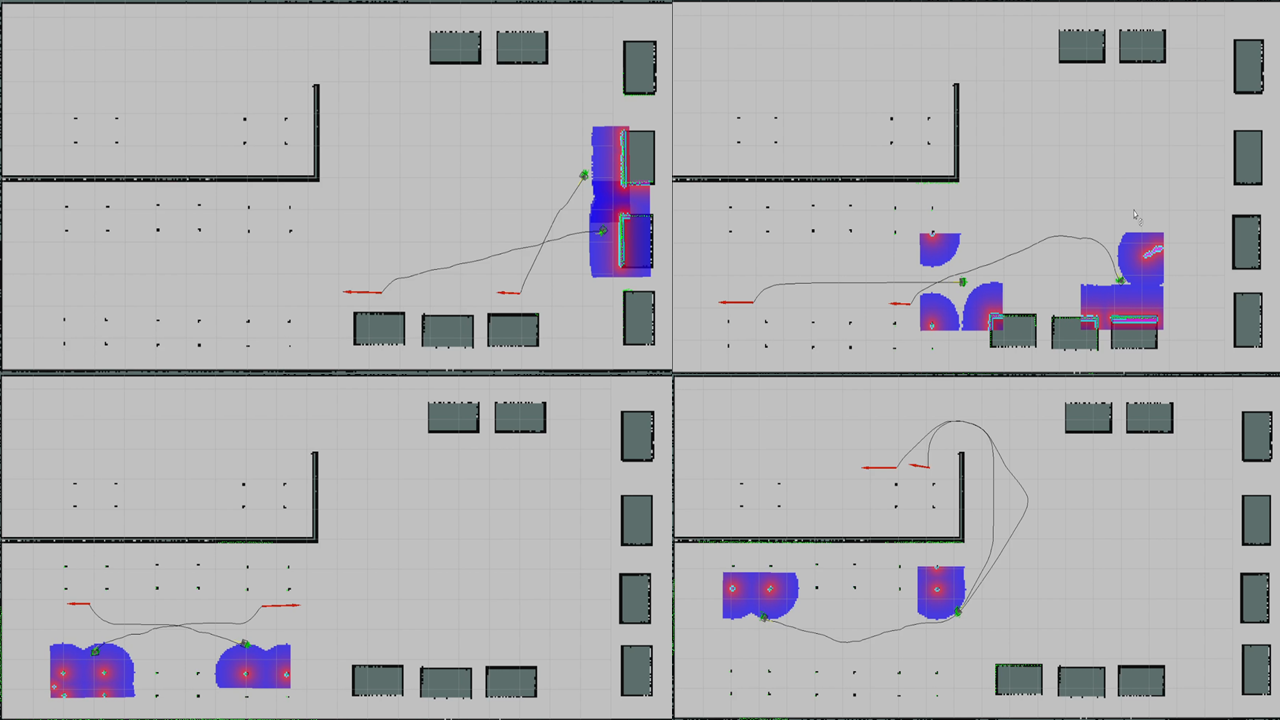}\vspace{-2mm}
	\caption{Multiple Turtlebot3 Robot performing different tasks}
	\label{fig:Multiple_Turtlebot3_tasks}
\end{figure}

\paragraph*{Simulation with two Turtlebot3 robots}
Fig. \ref{fig:Multiple_Turtlebot3_simulation} and Fig. \ref{fig:Multiple_Turtlebot3_tasks} showcase the tasks executed by the multiple turtlebot3, namely turtlebot\_01 and turtlebot\_02. In Fig. \ref{fig:Multiple_Turtlebot3_simulation}, both robots move from the infeed buffer to the Milling workstations M1 and M3, respectively. Fig. \ref{fig:Multiple_Turtlebot3_tasks} displays the robots' movements during task execution, with the top left panel illustrating the movement of both Turtlebot3s from the Grinding stations M4 and M5 to the CMM stations M7 and M9. The top right panel shows their movement from the CMM stations M7 and M9 to the Marking stations T4 and T6. The bottom left panel depicts the robots' movement from the Marking stations T4 and T6 to the Inspection Stations T1 and T3, and the bottom right panel demonstrates the robots delivering the finished product to the outfeed buffer after completing the task at the Marking stations. The simulation results with multiple Turtlebot3 robots using the adaptive goal management system are shown in the video\footnote{\url{https://youtu.be/LXhjw1-pLto}}.

\section{Conclusion and Future Work}
This paper proposed a novel cloud-based robot management system, AGM, for operating industrial robots in various application domains. The AGM system uses an adaptive goal execution approach and restful API communication to efficiently allocate and execute multiple tasks. The simulations using {\tt ROS1} with {\tt Gazebo} demonstrate the effectiveness and scalability of the AGM system. The future work will be to extend the AGM system's capabilities by incorporating optimization, machine learning and artificial intelligence techniques, exploring the use of swarm intelligence algorithms, and investigating the use of edge computing to enable real-time decision-making. We will also focus on testing the AGM system in different industrial domains to explore and demonstrate its effectiveness in various real-world applications. This research work emphasized the potential of the proposed cloud-based AGM system and highlighted the need for further research to enhance its capabilities and effectiveness of managing robots to meet the different industrial requirements.

\section*{Acknowledgment}
This research was supported by the Basic Science Research Program through the National Research Foundation of Korea (NRF) funded by the Ministry of Education (NRF- 2020R1F1A1076404, NRF-2022R1F1A1076260).

\bibliographystyle{ieeetr}
\bibliography{AGM}

\end{document}